\documentclass[sigconf]{acmart}
\PassOptionsToPackage{bookmarks={false}}{hyperref}
\AtBeginDocument{%
  \providecommand\BibTeX{{%
    \normalfont B\kern-0.5em{\scshape i\kern-0.25em b}\kern-0.8em\TeX}}}

\copyrightyear{2020}
\acmYear{2020}
\acmConference[MOBILESoft '20]{IEEE/ACM 7th International Conference on Mobile Software Engineering and Systems}{October 5--6, 2020}{Seoul, Republic of Korea}
\acmBooktitle{IEEE/ACM 7th International Conference on Mobile Software Engineering and Systems (MOBILESoft '20), October 5--6, 2020, Seoul, Republic of Korea}
\acmPrice{}
\acmDOI{}
\acmISBN{}

\begin{document}

%%
%% The "title" command has an optional parameter,
%% allowing the author to define a "short title" to be used in page headers.
\title[ReviewViz: Assisting Developers Perform Empirical Study]{ReviewViz: Assisting Developers Perform Empirical Study on Energy Consumption Related Reviews for Mobile Applications}

%%
%% The "author" command and its associated commands are used to define
%% the authors and their affiliations.
%% Of note is the shared affiliation of the first two authors, and the
%% "authornote" and "authornotemark" commands
%% used to denote shared contribution to the research.

\author{Mohammad Abdul Hadi}
\affiliation{%
  \institution{University of British Columbia}
  \streetaddress{3187, University Way}
  \city{Kelowna}
  \country{Canada}}
\email{hadi@alumni.ubc.ca}

\author{Fatemeh Hendijani Fard}
\affiliation{%
  \institution{University of British Columbia}
  \streetaddress{3333, University Way}
  \city{Kelowna}
  \country{Canada}}
\email{fatemeh.fard@ubc.ca}

%%
%% The abstract is a short summary of the work to be presented in the
%% article.
\begin{abstract}
    
    Improving energy efficiency of mobile applications is a topic that has gained a lot of attention recently. It has been addressed in a number of ways such as identifying energy bugs and developing a catalog of energy patterns. Previous work shows that users discuss the battery related issues (energy inefficiency or energy consumption) of the apps in their reviews. However, there is no work that addresses the automatic extraction of the battery related issues from users' feedback. 
    
    In this paper, we report on a visualization tool that is developed to empirically study machine learning algorithms and text features to automatically identify the energy consumption specific reviews with the highest accuracy. Other than the common machine learning algorithms, we utilize deep learning models with different word embeddings to compare the results. 
    Furthermore, to help the developers extract the main topics that are discussed in the reviews, two state of the art topic modeling algorithms are applied. The visualizations of the topics represent the keywords that are extracted for each topic along with a comparison with the results of \textit{string matching}. 
    
    The developed web-browser based interactive visualization tool is a novel framework developed with the intention of giving the app developers insights about running time and accuracy of machine learning and deep learning models as well as extracted topics. 
    The tool makes it easier for the developers to traverse through the extensive result set generated by the text classification and topic modeling algorithms. The dynamic-data structure used for the tool stores the baseline-results of the discussed approaches and are updated when applied on new datasets. The tool is open sourced to replicate the research results. \footnote{https://github.com/Mohammad-Abdul-Hadi/Empirical\_Study\_on\_Text\_Classification} 
    
\end{abstract}

%%
%% The code below is generated by the tool at http://dl.acm.org/ccs.cfm.
%% Please copy and paste the code instead of the example below.
%%

%%
%% Keywords. The author(s) should pick words that accurately describe
%% the work being presented. Separate the keywords with commas.
\vspace{-4mm}

%% A "teaser" image appears between the author and affiliation
%% information and the body of the document, and typically spans the
%% page.
\begin{teaserfigure}
  \centering
  \includegraphics[width=\textwidth]{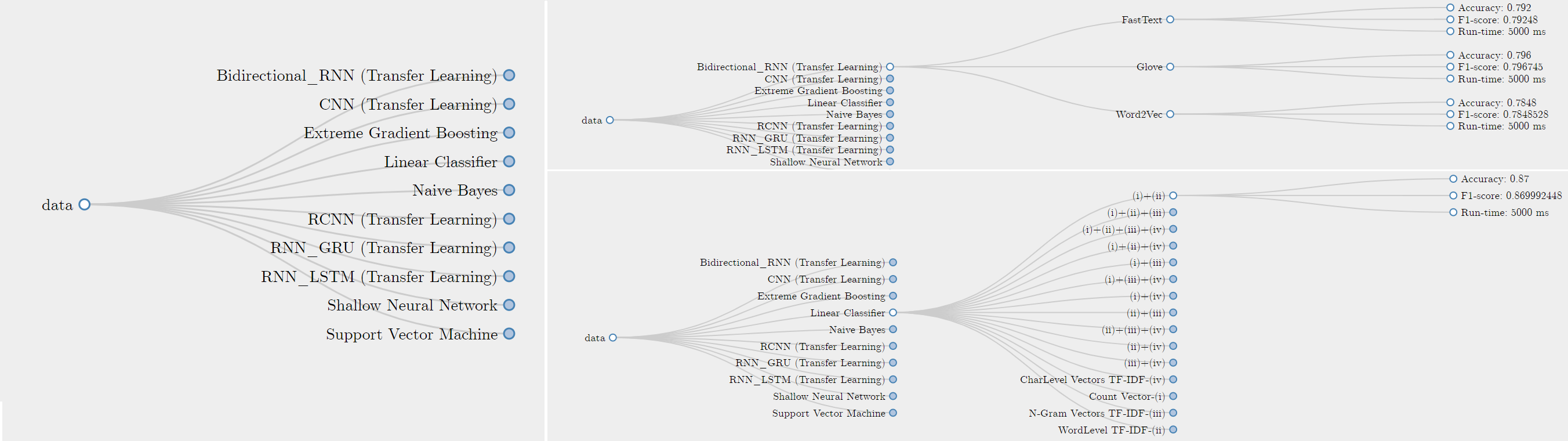}
  \caption{Visualization tool for the easy traversal of result-set produced by the elaborate empirical study}
  \Description{Teaser Figure}
  \label{fig:teaser}
\end{teaserfigure}

%%
%% This command processes the author and affiliation and title
%% information and builds the first part of the formatted document.
\maketitle
\vspace{-2mm}

\section{Introduction}

\begin{figure*}[t]
  \includegraphics[width=.9\linewidth]{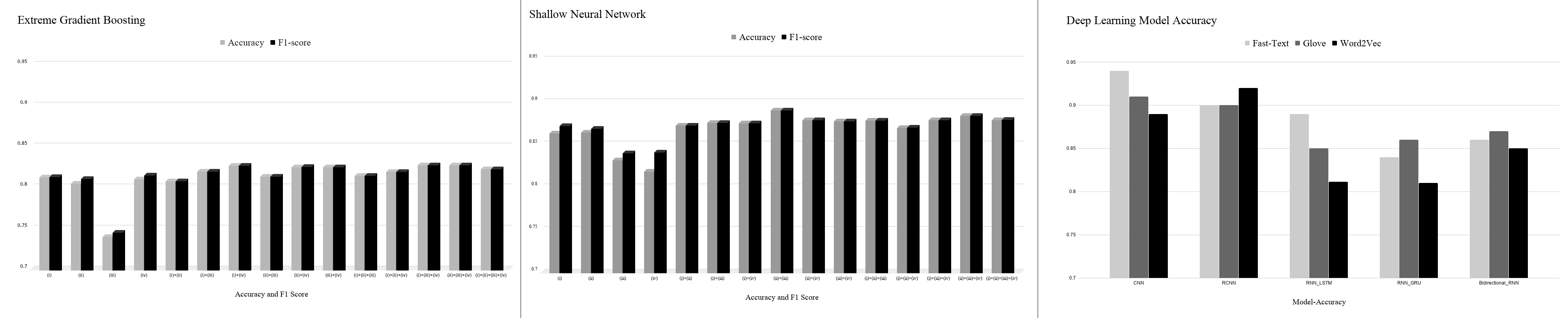}
  \caption{Different Illustrations of the Empirical Data}
  \label{fig:totImg}
\end{figure*}

Improving energy efficiency of mobile applications is a topic that has gained a lot of attention recently \cite{cruz2019catalog}. This challenge is addressed in a number of ways including energy consumption profiling \cite{chowdhury2019greenscaler}, developing a catalog of energy patterns \cite{cruz2019catalog}, and investigating users' reviews discussed on app stores \cite{wilke2013energy}. Previous work shows that users discuss the battery related issues (energy inefficiency or energy consumption) of the apps in their reviews \cite{wilke2013energy}, \cite{noei2019too}. However, there is no work that addresses the automatic extraction of the battery related issues from users' feedback. Although many works classify the app users' reviews automatically \cite{chen2014ar}, there is still a need to retrieve reviews on specific features \cite{dkabrowski2019finding}.

Considering the importance of energy efficiency in mobile apps \cite{cruz2019catalog} and the fact that it is a main topic discussed by app users \cite{noei2019too}, we empirically studied 90 different models to extract energy-related issues from app reviews. These models include different machine learning (with various feature combinations) and deep learning models. 
The usage of deep learning models leveraged with three pre-trained word embeddings is a first step toward transferring the knowledge for automatic extraction of specific features from app users' reviews. 

The results of computational analysis can be effectively combined with human reasoning with the help of powerful visualization and can lead to increased efficiency and more valuable results \cite{infoviz}.
For the empirical study, we developed an interactive web based visualization tool that enables the users compare the accuracy of the results and the computational time of different algorithms, as well as investigating the main topics in the filtered reviews.
We developed the tool for the usage of app developers in mind, giving them the flexibility of choosing among different techniques based on their requirements for running time or accuracy. 

In this paper, we explain the architecture and technical details of the developed tool. 
The tool is evaluated by 8 participants and is effective in examining the results set, when exploring a large number of models, feature combinations, and techniques. 
The main contributions of our work are empirical studies of machine learning and deep learning models for automatic extraction of app-specific features (energy consumption) from user reviews, and developing a visualization tool for app developers to explore the results. 

\vspace{-4mm}
\section{Tool Architecture}

For developing the tool, we have assessed different methods \cite{four} to find a convenient and concise way of presenting the generated data so that developers can compare and utilize the extracted information with maximum efficiency.
The tool contains two main components that are structured based on two essential purposes.
%(i) Present the extensive outcome of classification techniques of the energy-related reviews and (ii) Demonstrate the outcome of topic modeling approaches to automatically discover underlying issues that cause energy inefficiency. 
Each component is necessary to accomplish insightful data interpretation for helping developers to process all the information in a consistently comparable way. This section discusses both components of the visualization tool.
\vspace{-4mm}
\subsection{Component to Visualize Results of Different Classification Techniques}
\label{sec:component1}
\begin{figure}[b!]
  \includegraphics[scale=0.25]{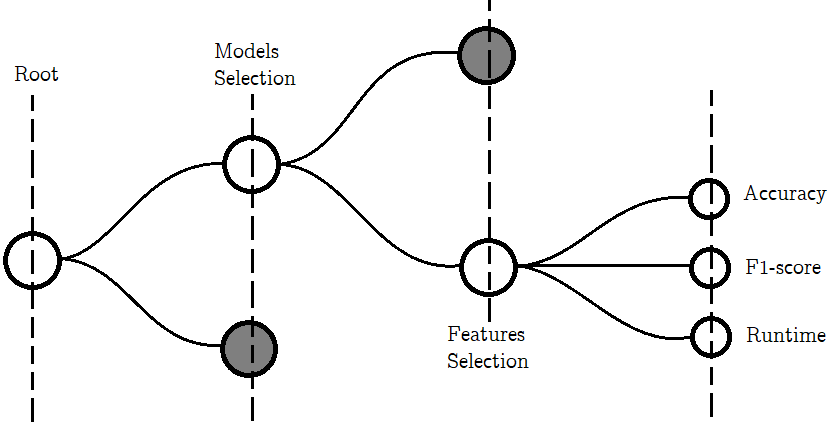}
  \caption{Analysis of visualization technique presented in Figure \ref{fig:teaser}}
  \label{fig:dissection}
\end{figure}

\begin{figure*}[t]
  \includegraphics[width=.8\linewidth, scale=0.7]{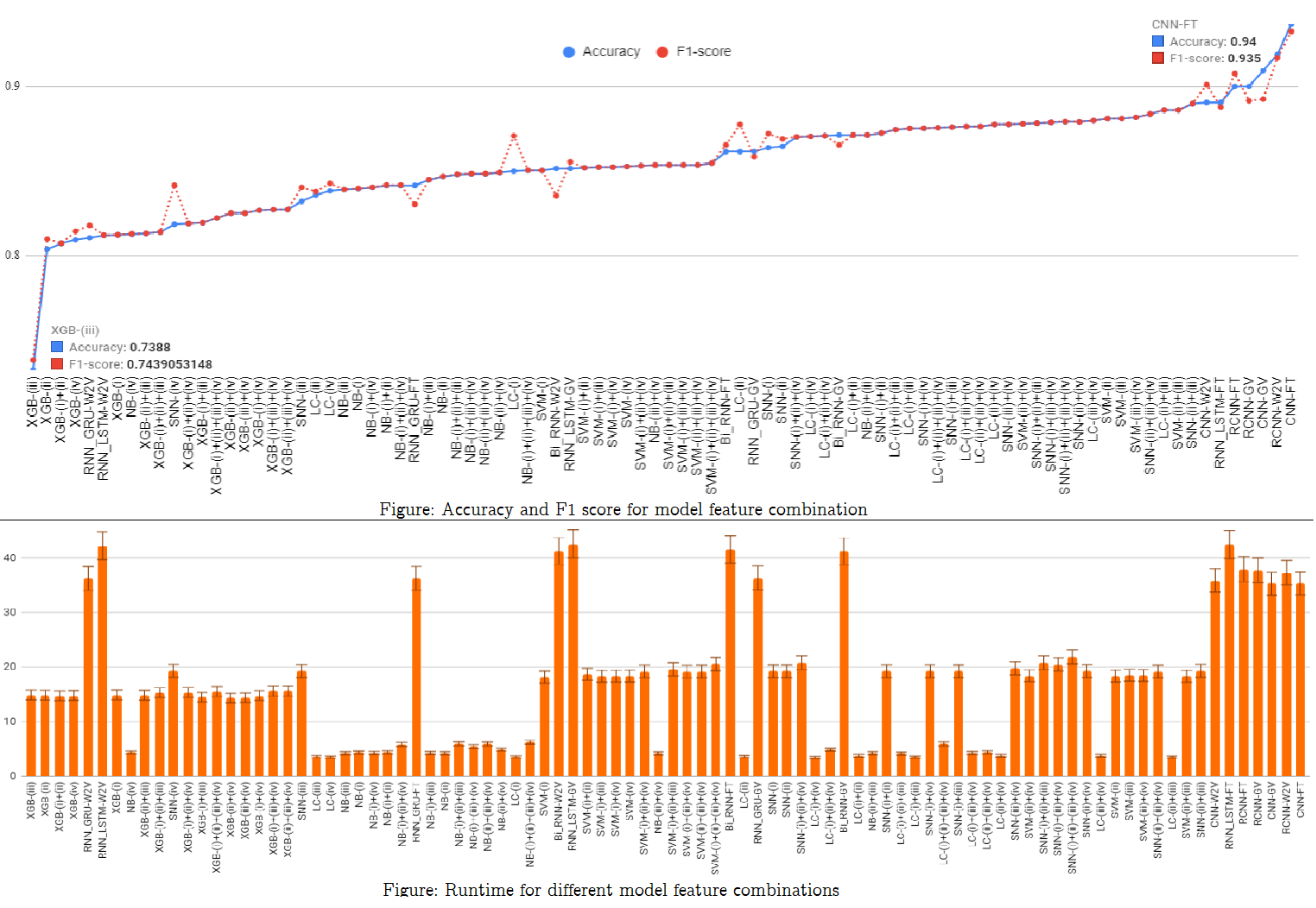}
  \caption{Different Illustrations (Overall Representation) of the Empirical Data}
  \label{fig:totImg2}
\end{figure*}

This component of the tool provides various visualizations: (1) a tree visualization that enables the users to explore the results by selecting a model, features of the model, and the evaluation metrics (e.g. accuracy, running time); (2) various plots such as bar chart and line chart that provides comparisons of the evaluation metrics for different feature combinations of the same algorithm (see Fig. \ref{fig:totImg}) and comparison of one or more evaluation metric among all of the trained models (see Fig. \ref{fig:totImg2}).

For the classification of energy-related data, four developers exhaustively performed 90 combinations of different models and features, recording the accuracy, F1-score, Precision, Recall, and Runtime information of each model. 
In Fig. \ref{fig:teaser}, we presented a comprehensive visualization technique for users' effortless traversal of the result set. This horizontal interactive visualization tree helps developers choose their itinerary without being exposed to the complete information. We have used a JavaScript library named Data-Driven Document (\textit{D3.js}), which helps to produce dynamic, interactive data visualizations in web browsers \cite{wikid3}. The reason behind choosing \textit{D3.js} is that this library uses a wide variety of pre-built JavaScript functions to create and manipulate \textit{Scalable Vector Graphics (SVG)} objects and apply suitable transitions and dynamic effects to them. Furthermore, large datasets of various formats, such as \textit{JSON} and CSV, can easily be bound to these objects to generate rich graphic charts and diagrams. Our visualization tool depends on the dataset compiled in \textit{JSON} file format. Using the data binding technique, we load the dataset from the generated file and use the data to drive the creation of elements such as an SVG object. 
%The properties and behaviors associated with the object depend on the values of attributes provided in the \textit{JSON} file.

The visualization tree consists of four levels. The dissection of the structure is shown in Fig. \ref{fig:dissection}. The levels are annotated as the \textit{Root}, \textit{Model Selection}, \textit{Feature Selection}, and \textit{Terminal}. The \textit{Root} works as the starting point for the user; when completely collapsed, only \textit{Root} node is visible. Upon clicking this \textit{Root} node, all the nodes (referencing to 10 different models) of the second horizontal level, \textit{Model Selection}, get expanded for user selection. If we add any new untrained model, the associated node for that model will appear in a different color. Each node has a dynamic tool-tip. The size of each node in the \textit{Model Selection level} is determined by the number of different features performed on the respective model. A higher number of features for a model would result in a bigger node.

A further selection of a node (model) from the second level would enforce all its' associated nodes form the third level (\textit{Feature Selection level}) to expand. These nodes represent features that are used to train the model (for performance analysis). Selecting any node from the third level would expand the three next-level nodes for \textit{Accuracy}, \textit{F1-score}, and \textit{Run-time}. Each node has tool tips to state the actual values of these metrics. These nodes also vary in sizes depending on the value of the metrics. 
%Only nodes referencing to F1-score can be further expanded into nodes referencing \textit{Precision} and \textit{Recall}.

%If a node is expandable into further nodes, that node is \textit{fill}ed with light blue color; and the non-expandable nodes are left transparent with a colored outline. 
All the nodes can be dragged, dropped, zoomed, and collapsed anywhere on the screen to provide more flexibility and convenience for the users. Zooming in can be performed by either double click or by scrolling the mouse-wheel; zooming out can be performed by holding shift when double-clicking. Clicking on the desired node achieves the expansion and collapse of the node. The tree auto-calculates its sizes both horizontally and vertically so that the maximum number of nodes can be presented on the screen while keeping the view comfortable and aesthetically pleasing.

Users can choose from a range of techniques to visualize the data; the options include horizontal, vertical bar-charts, line chart, 
%a combination of line and column chart, 
and scatter plot. For brevity, we chose to include only the outputs of horizontal column charts in Fig. \ref{fig:totImg}. Here, all traditional machine learning models have individual charts, whereas the neural network models are presented together for better comparison purposes (right side of Fig. \ref{fig:totImg}). Each column of a chart represents a feature utilized with the respective model; the first and seconds bars of each column represent the accuracy and F1-score for the model-feature combination, respectively. Users can sort each chart in the ascending or descending order. 
%We found that users could not comprehend the complete information quickly as the result-set is split into pieces (models) for precise visualization. 
We modified the tool to give users the option to combine all these representations for any given metrics: accuracy, precision, recall, f1-score, and run-time. 
%Users can overlap visualizations of the same category (line chart or bar chart) for different metrics as long as the metrics use the same units in \textit{x} axis and \textit{y} axis. 
Top and bottom parts of Fig. \ref{fig:totImg2} demonstrate accuracy, f1-score combination and run-time for all model feature combinations respectively.

The \textit{.json} file used for data-binding in our visualization tool contains the average values for all metrics. Whenever a user performs any model-feature combination using the replication package of our empirical study, all the resulting information gets updated in the file, which in turn is loaded by the tool for data binding. Details of different parameterization of provided models are saved and the tool-tips display the changing outcome for user-convenience.
\vspace{-3mm}
\subsection{Component to Visualize Results of Different Topic Modeling Techniques}
\label{sec:component2}
\begin{figure}[ht]
  \includegraphics[scale=0.25]{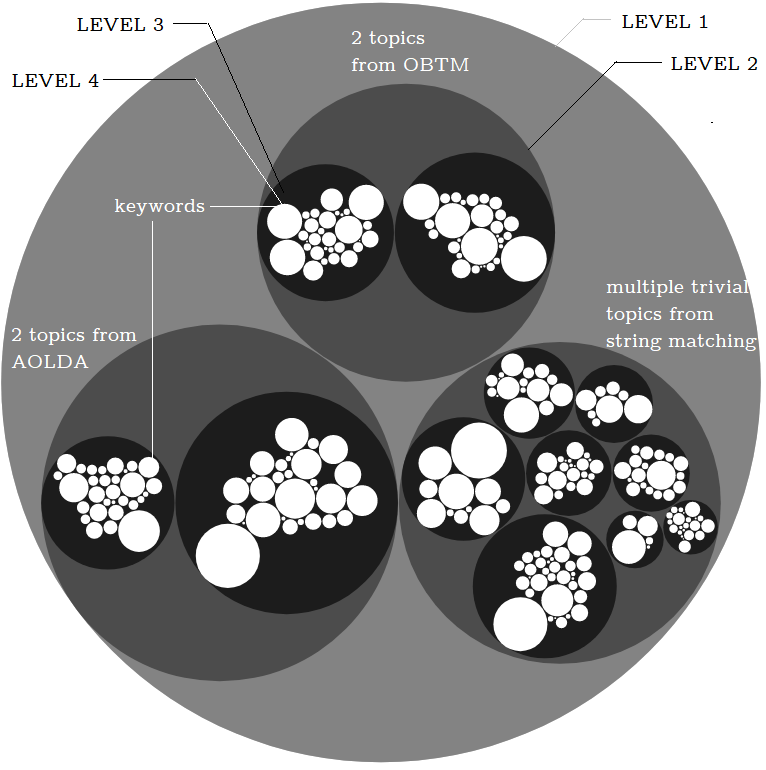}
  \caption{Circle Packing Visualization technique for keywords presentation in topic modeling}
  \label{fig:tm}
\end{figure}
In our study, we compared two state of the art topic modeling algorithms, namely \textit{OBTM} \cite{OBTM} and \textit{AOLDA} \cite{gao2018online} with the results of \textit{string matching} to automatically investigate the emerging issues responsible for energy inefficiency of the apps. OBTM is used as it can extract topics from short texts \cite{OBTM} and AOLDA is chosen because it can capture version-sensitive emerging topics from mobile app reviews \cite{gao2018online}. We used D3.js to develop the visualization tool shown in Fig. \ref{fig:tm} to exhibit the result of a qualitative study on the performance of the techniques. 
The result of a topic modeling algorithm is a number of topics with a set of keywords for each extracted topic, which are shown in the visualization. 

The topic visualization is divided into four levels: light gray (level-1), dark gray (level-2), black (level-3), and white (level-4) circle. The level-1 circle is the main container holding the results of all algorithms, where each one is represented in dark gray circles. The extracted topics for each algorithm are shown as black level-3 circles (one for each topic) with keywords inside them (level-4). The size of the white circles depends on the probability of the term (keyword) to appear in a review providing that the review falls into the term's associated topic. For string matching, the size of the circle depends on term's frequency in the review-set.

For example, in Fig. \ref{fig:tm}, OBTM has two main topics that are shown by black circles, with all of their keywords inside them respectively. 
In our study, the developers explored the dataset with string matching which yielded to only 2 main topics that are related to energy efficiency of the apps. Therefore, the number of topics (that should be set manually) for OBTM and AOLDA is set to two. 

% In this visualization, all the circles are zoomable and equipped with dynamic tool-tips and textual information. For further investigation, circles can be expanded to occupy the whole screen upon clicking. It helps the user to annotate each topic adequately based on the terms, as these keywords are presented with different sizes to represent their contribution in the extracted topic. Users can change visualization technique to "\textit{StreamRiver} visualization" which is excluded from the paper for brevity. 

\vspace{-3mm}
\subsection{Replication package}
\label{sec:rp}

To  encourage  replicability of the research results, we make all scripts, codes, and graphs available to the community \footnote{https://github.com/Mohammad-Abdul-Hadi/Result-Set-Visualization-d3js} and will provide the annotated dataset upon request.

\vspace{-3mm}
\section{Experiments}
\label{sec:rs}
We experimented the models on a well curated dataset of Google Play reviews. The dataset contains more than 400 apps from 24 different  app categories. The training data includes <date>, <rating>, <review>, <app\_name>, <app\_category> records that are labeled as energy related reviews. 
The results of our empirical study reveals that six deep learning models achieved the highest accuracy among 90 different model-feature combinations tested. However, as expected, the machine learning algorithms prevail when we take run-time into consideration. 
Due to the space limits, we report on the experiments and results of the study briefly. More details are provided in the tool page. 

Four developers participated in our empirical study who had the independence of not choosing the visualization tool for careful inspection of the dataset and models. However, all of them found that the visualization tool was effective for the exploration of the result set. Moreover, we invited four undergraduate students (who had the basic knowledge of machine learning and were paid) to evaluate the tool. 
The purpose of the tool development and the results were explained to the participants. They worked with the tool to explore the results and choose the models.  Specifically, we asked (i) whether the tool was necessary for exploring the results over a manual analysis, (ii) comprehensiveness of the visualizations, (iii) which (sub)-components that are not necessary for our purpose, and (iv) the sections of the tool that requires improvement. 

All of the participants reported that the visualization tool is required for exploring and deciding the model to use for filtering the reviews. One of the participants pointed out the lack of options presented in the topic modeling part, which we found to be true.
The comprehensiveness, smooth interactability, aesthetic features, and user-friendliness of the tool was also highly commended. 
\vspace{-3mm}
\section{Conclusion}

We have successfully delivered an interactive visualization tool to support the developers who frequently analyze and inspect users' feedback to identify energy efficiency issues. The tool is provided as part of the research on empirical studies of automatic extraction of app-specific features from app users' reviews with transfer learning. For this study, we focused on energy efficiency issues. 
In the future, we plan to work on providing different choices for result-set visualization of different topic modeling algorithms and perform an extensive qualitative study on the usage of the tool.
\vspace{-2mm}
%% The next two lines define the bibliography style to be used, and
%% the bibliography file.
\bibliographystyle{ACM-Reference-Format}
\bibliography{hadi-final-FHF}

\end{document}